\documentclass[journal=jacsat,manuscript=article]{achemso}

\usepackage{chemformula} 
\usepackage[T1]{fontenc} 
\usepackage{amsfonts} 

\usepackage{kotex}

\author{Jaechang Lim}
\affiliation[KAIST]
{Department of Chemistry, KAIST, Daejeon, South Korea}
\author{Seongok Ryu}
\affiliation[KAIST]
{Department of Chemistry, KAIST, Daejeon, South Korea}
\author{Kyubyong Park }
\affiliation[Kakao Brain]
{Kakao Brain, Pangyo, South Korea}
\author{Yo Joong Choe}
\affiliation[Kakao]
{Kakao, Pangyo, South Korea}
\author{Jiyeon Ham}
\affiliation[Kakao Brain]
{Kakao Brain, Pangyo, South Korea}
\author{Woo Youn Kim}
\affiliation[KAIST]
{Department of Chemistry, KAIST, Daejeon, South Korea}
\alsoaffiliation[Second University]
{KI for Artificial Intelligence, KAIST, Daejeon, South Korea}
\email{wooyoun@kaist.ac.kr}

\title[An \textsf{achemso} demo]
  {Predicting drug-target interaction using 3D structure-embedded graph representations from graph neural networks}

\abbreviations{IR,NMR,UV}

\begin{document}






\begin{abstract}
Accurate prediction of drug-target interaction (DTI) is essential for \textit{in silico} drug design. For the purpose, we propose a novel approach for predicting DTI using a GNN that directly incorporates the 3D structure of a protein-ligand complex. We also apply a distance-aware graph attention algorithm with gate augmentation to increase the performance of our model. As a result, our model shows better performance than docking and other deep learning methods for both virtual screening and pose prediction. In addition, our model can reproduce the natural population distribution of active molecules and inactive molecules.
\end{abstract}

\section{Introduction}
Accurate prediction of drug-target interaction (DTI) is essential for \textit{in silico} drug discovery.
Thanks to the revolutionary development of theories and the required computational power, calculation methods---such as molecular dynamics and quantum mechanics/molecular mechanics---show reliable performance for calculating the binding free energy between a ligand and a protein.\cite{Wang2017,Beierlein2011}
Though the accuracy of such methods is close to chemical accuracy, their huge computational costs impede their routine usage. On this ground, molecular docking has been developed to predict the binding affinity with affordable computational costs and reliable accuracy.\cite{Venkatachalam2003,Allen2015,Ruiz-Carmona2014,Zhao2013,Jain2003,Jones1997,Friesner2004,Trott2009} 
Molecular docking greatly reduces computational costs through principled parameter fitting, and as a result, it is fast enough to be practically used. 
Despite this advantage, however, the accuracy of docking is often insufficient to find novel drug candidates from huge molecular libraries. 

More recently, deep learning techniques\cite{LeCun2015} have attracted much attention as a promising alternative for calculating DTI. 
While various machine learning techniques have already been utilized to improve the performance of DTI calculation, deep learning techniques in particular have clear advantages over conventional approaches.
Deep learning techniques can automatically extract task-related features from a huge database without handcrafted features or rules. 
In addition, the huge expressive power of deep learning techniques enables efficient training with a large amount of data.
Such advantages are particularly important in, for instance, biological and chemical applications, where the sizes of the databases are exponentially growing.\cite{Park2019}


Wallach et al. were first to introduce a deep learning based method for predicting DTI.\cite{Wallach} 
In their work, the atomic coordinates of a protein-ligand complex around the binding site are represented on a 3D rectangular grid.
Then, a 3D convolutional neural network is applied to the rectangular grid to classify whether the complex is active or not. 
As a result, their model outperformed docking in terms of the area under the receiver operating characteristic (AUROC) and the adjusted LogAUC\cite{Mysinger2010}. Ragoza et al. used a similar neural network structure to extend the above approach to classify not only active and inactive ligands but also active and inactive binding poses.\cite{Ragoza2017}
For both classification tasks, the 3D convolutional neural network outperformed docking, RF-score\cite{Ballester2010}, and NNScore\cite{Durrant2011}. 
Also, 3D deep convolutional neural networks can improve the accuracy of predicting absolute binding affinities\cite{Jimenez2018,Stepniewska-Dziubinska2018} compared to classical scoring functions such as RF-Score\cite{Ballester2010}, X-Score\cite{Wang2002}, and Cyscore\cite{Cao2014}. 

Another set of approaches independently expresses a ligand and a protein without 3D binding structures between them. A common framework is to integrate the vectors representing each of the ligand and the protein within a neural network such as a siamese network. 
The main advantage of these approaches is that they do not require binding poses, whose high-quality data is limited. 

Karimi et al. developed a method named DeepAffinity to predict the pIC50 of a given protein-ligand complex with only the sequence of the protein and the SMILES of the ligand.\cite{Karimi2018} 
They also interpreted how their model treats the interaction between ligands and proteins using an autoencoder, which is jointly trained with the prediction model. 
In spite of the advantage that their model doesn't require 3D geometry data, the performance significantly dropped for the test set where both ligands and proteins were not included in the training set. 
Gao et al. developed a method for predicting DTI based on a siamese network composed of a graph convolutional neural network and a recurrent neural network.\cite{Gao} They analyzed how their model learns protein-ligand interactions with attention algorithm. 
Ozt\"urk et al.\cite{Ozturk2018} and Lee et al.\cite{Lee2018} proposed a method based on 1D deep convolutional neural networks with representing a ligand as a SMILES and a protein as a sequence.
Additionally, it is reported that a neural fingerprint specialized in DTI prediction can increase the accuracy of DTI models compared to conventional fingerprints.\cite{Gonczarek2016}

Accurately representing protein-ligand complexes is critical for the performance of DTI models and determines the corresponding neural network architecture. 
Furthermore, the representation of protein-ligand complexes determines which dataset should be used and how to preprocess it. Graph representation of molecules, where atoms and chemical bonds correspond to nodes and edges respectively, is a natural and compact way of describing molecular structures. Sophisticated graph neural networks (GNNs)\cite{Gilmer2017,Li2015,Duvenaud2015,Kearnes2016,Battaglia2016,Schutt2017} with graph representation of molecules showed remarkable performance for predicting various molecular properties.

The success of GNNs in predicting molecular properties suggests the graph neural network as a promising architecture for improving DTI models.  
Indeed, Gomes et al. developed an atomic convolutional neural network (ACNN) by defining the atom type convolution and the radial pooling layer.\cite{Gomes2017} 
Feinberg et al. proposed a spatial graph convolution, applying different graph convolution filters based on the Euclidean distances between atoms.\cite{Feinberg2018} 
Torng et al. developed a two-step graph convolutional framework for predicting DTI.\cite{Torng2018} 
Motivated by these recent results, we further develop ways to improve GNN based DTI models. One way of achieving this is to devise a graph representation of 3D binding structures for correctly capturing intermolecular interactions. In accordance with the new representation, a corresponding neural network architecture should be designed.

In this paper, we propose a novel approach for predicting DTI using a GNN that directly incorporates the 3D structure of a protein-ligand complex.
This allows us to not only classify active and inactive compounds but also distinguish active and inactive binding poses. 
No heuristic chemical rules were used to deal with noncovalent interactions. 
Our GNN also includes a graph attention mechanism \cite{Velickovic2017}, which allows us to model the intrinsic nature of a protein-ligand complex: specifically, interactions between ligand atoms and protein atoms do not contribute equally to their binding affinity. 
Additionally, we improve the performance of our model by adopting a gate augmentation mechanism.\cite{Ryu2018} As a result, our method outperformed docking and previous deep learning methods in terms of both virtual screening and pose prediction. We also show that our model can reproduce the natural population distribution of active and inactive molecules. 
Finally, we analyzed the limitations of our model's generalization ability using an external molecular library.

\section{Method}
In terms of methodology, our contribution can be summarized in three parts: embedding the 3D structure of a protein-ligand complex in an adjacency matrix, devising attention algorithm considering 3D structures, and introducing a variant of graph neural networks that is suitable for learning protein-ligand interaction. Each part will be described in the following subsections. Before explaining our contributions, we will briefly introduce graph neural networks. Finally, we will explain which datasets are used and how to preprocess them.

\subsection{Graph neural network}
Graphs can be defined by ($V$, $E$, $\mathbf{A}$), where \textit{V} is a set of nodes, \textit{E} is a set of edges, and \textit{A} is an adjacency matrix. 
In an attributed graph, the attribute of each node is usually represented by a vector.
The adjacency matrix is an $N\times N$ matrix, where $\mathbf{A}_{ij}>0$ if $i$-th and $j$-th nodes are connected and $\mathbf{A}_{ij}=0$ otherwise. \textit{N} denotes the number of nodes in the graph. Graph neural networks (GNNs) which operate on graphs have been explored in a diverse range of domains and have shown remarkable performance in various applications.\cite{Zhou2018} Also, various architectures of GNNs have been developed.\cite{Gilmer2017,Li2015,Duvenaud2015,Kearnes2016,Battaglia2016,Schutt2017}

In general, GNNs are composed of three stages: i) updating node features, ii) aggregating the node features and processing graph features, and iii) predicting a label of the graph.\cite{Battaglia2018} In the first stage, the node feature $x_i$, representing attributes of the $i$-th node, is updated over several times of message passing between neighboring nodes. This stage aims to obtain high level representations of node features.
Then, the updated node features are aggregated to produce graph features. 
Here, the result of the aggregation must be invariant over permutations of node ordering.
Finally, the graph features are used to predict a label of the entire graph, for instance molecular properties. 

\subsection{Embedding the 3D structure of a protein-ligand complex}

To correctly predict drug-target interactions (DTIs), it is important to describe the 3D binding structure of a protein-ligand complex in a graph. 
The reason is that the 3D binding structure determines physical, chemical, and biological properties of complexes including intermolecular interactions. 
We embed the distance information between protein and ligand atoms in two adjacency matrices, $\mathbf{A}^1$ and $\mathbf{A}^2$. 
$\mathbf{A}^1$ represents purely covalent interactions, and $\mathbf{A}^2$ represents both covalent interactions and noncovalent intermolecular interactions. 
By constructing two adjacency matrices, we let our model learn how protein-ligand interactions affect the node feature of each atom. 
$\mathbf{A}^1$ and $\mathbf{A}^2$ are constructed as follows:
\begin{align}
  \mathbf{A}^1_{ij} & =
    \begin{cases}
      1 & \text{if $i$ and $j$ is connected by covalent bond or $i$=$j$}\\
      0 & \text{otherwise}\\
    \end{cases}
    \label{eq:A1ij}\\       
  \mathbf{A}^2_{ij} & =
    \begin{cases}
      \mathbf{A}^1_{ij} & \text{if $i,j$ $\in$ protein atoms or $i,j$ $\in$ ligand atoms}\\
      e^{-(d_{ij}-\mu)^2/\sigma} & \parbox[t]{.5\textwidth}{if $d_{ij}<5$ and $i\in$ ligand atoms and $j\in$ protein atoms, or if $d_{ij}<5$ and $i\in$ protein atoms and $j\in$ ligand atoms}\\[5ex]
      0 & \text{otherwise}\\
    \end{cases}
    \label{eq:A2ij}
\end{align}
where $d_{ij}$ is the distance between the $i$-th and $j$-th atoms, and $\mu$ and $\sigma$ are learnable parameters. 
The form $e^{-(d_{ij}-\mu)^2/\sigma}$ in Eq.~\ref{eq:A2ij} reflects that intermolecular bonds are weaker than covalent bonds and that their strengths are getting weaker as the bond distance increases. 

Representing 3D structures in an adjacency matrix has advantages over grid representation. In grid representation, a large amount of empty grid points can cause unnecessary computation and memory usages. Also, grid representation can lose distance information between atoms depending on the grid spacing. Furthermore, because it is not rotationally invariant, rotating atomic coordinates changes the prediction value of binding affinity. In contrast, our graph representation is compact and rotationally invariant. In addition, it enables efficient expression of the exact distance between atoms.

\subsection{A distance-aware graph attention mechanism}

The inputs of our graph attention layer are adjacency matrix, $A$, and the set of node features, $\textbf{x}^{in}=\left\{x_1^{in},x_2^{in},...,x_N^{in}\right\}$ with $x\in \mathbb{R}^{F}$, 
where $N$ is the number of nodes (i.e the number of atoms). $F$ is the dimension of the node feature. 
The graph attention layer produces a new set of node features $\textbf{x}^{out}=\left\{x_1^{out},x_2^{out},...,x_N^{out}\right\}$ with $x\in \mathbb{R}^{F}$. 
In order to achieve sufficient expressive power, each node feature is transformed by a learnable weight matrix $\textbf{W} \in \mathbb{R}^{F \times F}$ such that $x_i' = \textbf{W} x_i^{in}$. 
Then, the attention coefficient, $e_{ij}$, is obtained as follow:
\begin{align}
     e_{ij}=x_i'^T\textbf{E}x_j'+x_j'^T\textbf{E}x_i',
\end{align}
where $\mathbf{E}\in \mathbb{R}^{F \times F}$ is also a learnable matrix. The attention coefficient, $e_{ij}$, represents the importance of the $j$-th node feature to the $i$-th node feature. We forced $e_{ij}=e_{ji}$ by summing $x_i'^T\textbf{E}x_j'$ and $x_j'^T\textbf{E}x_i'$. 
To reflect graph structure, the attention coefficient, $e_{ij}$, is computed only for $j\in N_i$, where $N_i$ is the neighborhood of the $i$-th node. 
We define $N_i$ as the set of nodes of which $A_{ij}>0$ because our adjacency matrix reflects both connectivity and the normalized distance. To manipulate the scale of the attention coefficients across nodes, the attention coefficient is normalized across neighbors.
Additionally, we multiply $A_{ij}$ to the normalized attention coefficients to reflect that a node with a shorter Euclidean distance is more likely to be important than the others. It can be considered as an inductive bias. Consequently, the normalized attention coefficient, $a_{ij}$, is given by
\begin{align}
     a_{ij}=\frac{\exp(e_{ij})}{\sum\nolimits_{j \in N_i} \exp(e_{ij})}A_{ij}.
\end{align}
After the normalized attention coefficient, $a_{ij}$, is obtained, each node feature is updated as a linear combination of the node features of the neighboring nodes with the normalized attention coefficient: 
\begin{align}
    x_i''=\sum\nolimits_{j \in N_i}a_{ij}x_j^{'}.
\end{align}

We also augment a gate mechanism to directly deliver information of the previous node features to the next layer. It is found that a gate augmentation algorithm significantly improves the performance of a model.\cite{Ryu2018}
We implement the output of our gated graph attention layer as a linear combination of $x'$ and $x''$ as follows:
\begin{align}
    x_i^{out}=z_ix_i'+(1-z_i)x_i''
\end{align}
with
\begin{align}
    z_i=\sigma (\textbf{U}(x_i\|x_i')+b),
\end{align}
where $\textbf{U} \in \mathbb{R}^{2F\times 1}$ is a learnable vector, $b$ is a learnable scalar value, $\sigma$ denotes a sigmoid activation function, and $(\cdot\|\cdot)$ is a concatenation of two vectors.
$z_i$ can be interpreted as how much information of input node features will be directly delivered to the next layer.

\subsection{Neural network architecture}
The inputs of our neural network are $\textbf{x}$, $\textbf{A}^1$, and $\textbf{A}^2$. 
Two new node features, $\textbf{x}^1$ and $\textbf{x}^2$, are produced by the gate augmented graph attention layer respectively with $\textbf{A}^1$ and $\textbf{A}^2$, i.e., $\textbf{x}^1=GAT(\textbf{x}, \textbf{A}^1)$ and $\textbf{x}^2=GAT(\textbf{x}, \textbf{A}^2)$, where $GAT$ is the gate augmented graph attention layer.
It should be noted that one gate augmented graph attention layer is shared when computing $\textbf{x}^1$ and $\textbf{x}^2$. 
The output node feature, $\textbf{x}^f$, is obtained by subtracting $\textbf{x}^1$ from $\textbf{x}^2$: 
\begin{align}
    \textbf{x}^{out}=\textbf{x}^2-\textbf{x}^1.
\end{align}
By subtracting the two node features, $\textbf{x}^2$ and $\textbf{x}^1$, we let our model learn the difference between the structure when a ligand binds to a protein and the structure when they are separated.
After the feature vectors are updated by several gate augmented graph attention layers, the feature vectors are summed into one vector representing the graph of the protein-ligand complex:
\begin{align}
    x^{graph}=\sum\limits_{i=1}^N x_i.
\end{align}
Finally, multi-layer perceptrons are applied to $x^{graph}$ to classify whether the protein-ligand complex or the binding pose is active or not. 
ReLU activation functions are used between the layers, and a sigmoid function is used after the last layer.

\subsection{Dataset preparation}
We used the DUD-E\cite{Mysinger2012} and PDBbind\cite{Liu2017} v.2018 datasets to train and test our model. 
72 proteins and 25 proteins in the DUD-E set were used to train and test the model, respectively.
To remove undesirable redundancy, we divided the dataset in a way that no protein is present both in the training set and in the test set.
The 3D binding structures of protein-ligand copmlexes were obtained from docking calculations.

The PDBbind dataset, which provides experimentally verified binding structures of protein-ligand complexes, was used to train our model to distinguish active and inactive binding poses. 
For each sample in the PDBbind dataset, we performed docking calculation to generate possible binding poses of the protein-ligand complex.
A generated pose was labeled as a positive sample if the root mean square deviation (RMSD) from its experimentally verified binding structure is less than 2 {\AA} and labeled as a negative sample if the RMSD is larger than 4 {\AA}. The samples whose RMSD is between 2 {\AA} to 4 {\AA} were omitted. 
Then, the PDBbind dataset was splitted into a training set and a test set depending on the protein so that the training and the test sets don't share common proteins. 
Additionally, the PDBbind samples whose protein is included in the DUD-E dataset were removed from both the training and the test sets. 

The statistics of our training and test sets is summarized in Table \ref{tab:Table1}. As Table \ref{tab:Table1} shows, the inactive samples and DUD-E samples are much more abundant in the training set and the test set compared to the active samples and PDBbind samples. To deal with such imbalance of the datasets, we sampled DUD-E active, DUD-E inactive, PDBbind positive, and PDBbind negative samples with the fixed ratio of 1:1:1:1 in constructing one training batch. 
\begin{table}
  \caption{Number of the training and the test samples for DUD-E active, DUD-E inactive, PDBbind positive, and PDDbind negative.}
  \begin{tabular}{lrrrr}
    \hline
      & DUD-E active  & DUD-E inactive & PDBbind positive & PDBbind negative \\
    \hline
    training   & 15864 &  973260& 1598 & 9511 \\
    test & 5841 & 364149 & 496 & 2735 \\
    \hline
  \end{tabular}
  \label{tab:Table1}
\end{table}

All the 3D binding structures of protein-ligand complexes were obtained using Smina\cite{Koes2013}, the fork of AutoDock Vina\cite{Trott2009}, even when experimental 3D structures are available, to maintain consistency. 
For the DUD-E dataset, the default setting of Smina was used. Exhaustiveness=50 and num modes=20 were used for docking calculations of the PDBbind dataset.
After the docking calculations were completed, the protein atoms whose minimum distance to the ligand atoms is larger than 8 {\AA} were excluded to remove unnecessary atoms in the graph representation of protein-ligand complexes.

We represent initial atom features as a vector of size 56. The 1st--28th entities represent ligand atoms and the 29th--56th entities represent protein atoms. 
The list of the initial atom features is summarized in Table  \ref{tab:Table2}.
Our model consists of four gate augmented graph attention layers and three fully connected layers.
The dimension of the graph attention layers was 140, and that of the fully connected layers was 128.
We trained our model for 150,000 iterations with the batch size of 32. 
To avoid overfitting, we applied dropout with the rate of 0.3 to every layer except the last of the fully connected layers.

\begin{table}
  \centering
  \caption{ The list of atom features }
  \begin{tabular}{llllll}
  \hline
    & atom type   & C, N, O, S, F, P, Cl, Br, B, H (onehot) & \\
    & degree of atom   & 0, 1, 2, 3, 4, 5 (onehot) & \\
    & number of hydrogen atoms attached & 0, 1, 2, 3, 4 (onehot) &\\
    & implicit valence electrons& 0, 1, 2, 3, 4, 5 (onehot) &\\
    & aromatic & 0 or 1 &\\
    \hline
  \end{tabular}
  \label{tab:Table2}
\end{table}

\section{Results and discussion}
 
\subsection{Performance on the DUD-E and PDBbind test set}
The performance of structure-based virtual screening can be assessed by measuring the ability to classify active and inactive compounds. 
We compared the performance of our models with docking and other deep learning based models in terms of AUROC, adjusted LogAUC, and PRAUC. 
We calculated the AUROC, adjusted LogAUC, and PRAUC of our model by averaging their values of each protein to balance data inequality between the proteins.

Table \ref{tab:Table3} summarizes AUROC, adjusted LogAUC, and PRAUC values of our model, docking, and other deep learning methods for the DUD-E test set. 
Among various deep learning models, we chose the deep learning models trained using the DUD-E dataset for fair comparison.
Though all the models used the one same dataset, we note that the exact division of the training and test sets is different over the models.
Table \ref{tab:Table3} clearly shows that our GNN-based method outperformed the other deep learning models as well as docking. Our model achieved AUROC of 0.968 compared to 0.689 of docking and 0.85--0.9 of other deep learning models. In addition, our model achieved high PRAUC value of 0.697 for the DUD-E test set, where the decoy molecules are much dominant than the active molecules.   

We also analyzed ROC enrichment (RE) \cite{Jain2008,Nicholls2008} score and summarized it in Table \ref{tab:Table4}. The RE score indicates the fraction of the true positive rate (TPR) to the false positive rate (FPR) at a certain FPR value. 
In terms of the RE score, our method is 5--6 times better than docking and 1.5--2 times better than other deep learning based models. Also, the distance-aware attention algorithm clearly improved the performance in virtual screening for all metrics.
Generally, in hit discovery, only top hundreds of molecules are selected and tested through experiments.
So the better performance on LogAUC and RE score indicates a practical advantage of our model when screening a large molecule set. 

\begin{table}
  \centering
  \caption{AUROC, adjusted LogAUC, and PRAUC of our model, docking, and other deep learning models. We note that the exact division of the training and test sets is different over the models.}
  \begin{tabular}{l|ccc}
    \hline
      & AUROC & adjusted LogAUC & PRAUC \\
    \hline
     ours & \textbf{0.968} & \textbf{0.633} & \textbf{0.697}   \\
     ours w/o attention & 0.936 & 0.577 & 0.623   \\
     docking   & 0.689 &  0.153 & 0.016   \\   
     Atomnet\cite{Wallach} & 0.855 & 0.321 & -   \\
     Ragoza et al.\cite{Ragoza2017} & 0.868 & - & -   \\
     Torng et al.\cite{Torng2018} & 0.886  & - & -   \\
     Gonczareket al.\cite{Gonczarek2016} & 0.904 & - & -   \\
    \hline
  \end{tabular}
  \label{tab:Table3}
\end{table}

\begin{table}
  \centering
  \caption{ ROC enrichment (RE) score  of our model, docking and other deep learning methods. The RE score indicates the fraction of the true positive rate (TPR) to the false positive rate (FPR) at a certain FPR value.}
  \begin{tabular}{lllll}
    \hline
    & 0.5\%  & 1.0\% & 2.0\% & 5.0\%  \\
    \hline
    ours   &  \textbf{60.431} &   \textbf{50.106} &  \textbf{33.465} & \textbf{16.713}  \\
    ours w/o attention  & 57.793 & 45.636 & 30.246 & 15.398  \\
    docking   & 9.307 &  7.151 & 5.380 & 3.803  \\
    Ragoza et al.\cite{Ragoza2017} & 42.559 & 29.654 & 19.363 & 10.710 \\
     Torng et al.\cite{Torng2018}& 44.406  & 29.748 & 19.408 & 10.735 \\
    \hline
  \end{tabular}
  \label{tab:Table4}
\end{table}

Distinguishing active and inactive binding poses between a ligand and a protein is important for interpreting their interaction. 
Such interpretation helps human experts rationally modify the ligand to improve its efficacy. Our model can be used for pose prediction because the 3D conformational information is directly included in the graph representation of a protein-ligand complex. 
Table \ref{tab:Table5} summarizes the performance of our model and docking in terms of AUROC and PRAUC for the PDBbind test set. 
Our model improves AUROC and PRAUC about 0.11 and 0.26 compared to docking respectively. As in the virtual screening results, the attention algorithm clearly improved the performance in pose prediction.

\begin{figure}[h!]
   \includegraphics[width=0.5\linewidth]{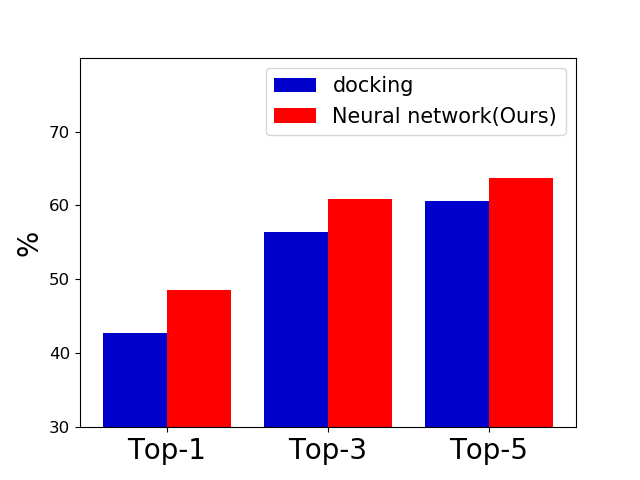}
   \caption{Percentage of complexes whose RMSD is lower than 2{\AA} identified in top-N poses by docking and our model}
 \label{fig:Figure1}
 \end{figure}

Figure \ref{fig:Figure1} shows the percentage of complexes with RMSD lower than 2{\AA} identified in top-N poses by docking and our model. 
Our model is 5\% to 7\% better than docking for the top-N pose prediction.  
However, the performance gap between our model and docking for top-N pose prediction is relatively small compared to the gaps in AUROC and PRAUC. 
This means that docking calculation is relatively accurate for ranking poses of different ligands with an identical protein.

\begin{table}
  \centering
  \caption{ AUROC and PRAUC of our model and docking for classification of active and inactive binding poses.}
  \begin{tabular}{lll}
    \hline
     & AUROC  & PRAUC \\
    \hline
    ours   &  \textbf{0.935} &   \textbf{0.772} \\
    ours w/o attention  &  0.927 &   0.698 \\
    docking   & 0.825 &  0.509 \\
    \hline
  \end{tabular}
  \label{tab:Table5}
\end{table}

\subsection{Distribution of predicted activity for a molecular library}

\begin{figure}[h!]
   \includegraphics[width=0.5\linewidth]{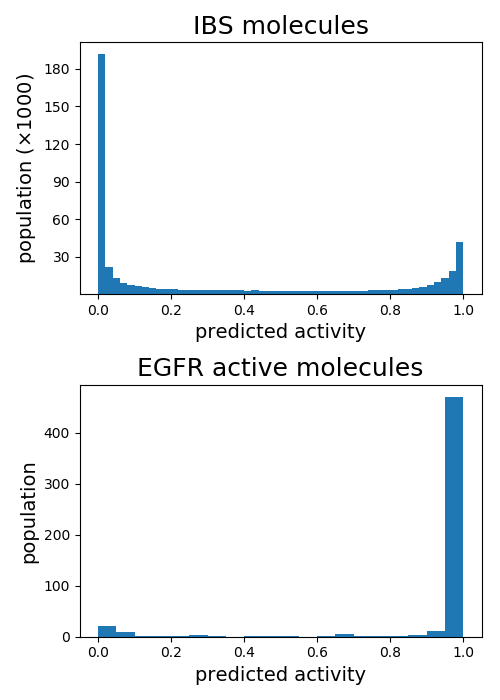}
   \caption{Activity distributions predicted by our model for the IBS molecules and EGFR active molecules in the DUD-E dataset.}
   \label{fig:Figure2}
 \end{figure}

The number of potential drug candidates is estimated about $10^{23}$ to $10^{60}$.\cite{Polishchuk2013} Among such large number of molecules, we can safely assume that most of them are inactive to a given protein. We tested whether our model can reproduce such natural activity distribution. 
To do so, 470094 synthetic compounds were obtained from https://www.ibscreen.com. We preprocessed those synthetic compounds (IBS molecules) as done for the DUD-E dataset. The distribution of predicted activities for the IBS molecules to epidermal growth factor receptor (EGFR) is plotted in Figure \ref{fig:Figure2}. It should be noted that EGFR is not included in the DUD-E training set. 

As shown in Figure \ref{fig:Figure2}A, our distribution has inactive molecules as a dominant portion, reproducing the natural population.
Also, in Figure \ref{fig:Figure2}B, our model correctly predicted the activity of most of the EGFR active molecules close to 1.0. The EGFR active molecules were obtained from the DUD-E dataset.
In the predicted activity distributions, a small peak is observed around 1.0 in Figure \ref{fig:Figure2}A and 0.0 in Figure \ref{fig:Figure2}B.
These unnatural peaks may come from overconfidence of our model, indicating a possibility of slight overfitting.

\subsection{Performance on external libraries: ChEMBL and MUV}
In the DUD-E dataset, the decoy molecules have considerable structural differences with the active molecules, so that classifying the DUD-E active and decoy molecules might be relatively easy.
On the other hand, experimentally verified inactive compounds share more common structures with active compounds than the decoys do.
Therefore, we further validated whether our model, trained on the DUD-E dataset, can classify active and inactive compounds which were experimentally verified.
The active and inactive molecules with respect to the DUD-E test proteins were collected from the ChEMBL\cite{Vanon2016} database and preprocessed in the same way as done for the DUD-E dataset.
We labeled the ChEMBL molecules whose IC50 is smaller than 1.0$\mu$ as active and inactive otherwise.
As a result, 27389 active and 26939 inactive molecules were obtained. 
Table \ref{tab:Table6} shows AUROC of our model and docking for the ChEMBL molecules. 
Though our model is still better than docking, the AUROC is dropped about 0.3 compared to the AUROC of the DUD-E test set. 
The ChEMBL molecules and the DUD-E test set share common proteins, so the difference in AUROC only comes from the difference between the ligand sets. 

\begin{table}
  \centering
  \caption{AUROC of our model, docking, and other deep learning models on experimentally verified active and inactive molecules, and MUV set. The experimentally verified active and inactive molecules were obtained from the ChEMBL database.}
  \begin{tabular}{lcc}
    \hline
      AUROC & ChEMBL  & MUV  \\
    \hline
    ours   &   \textbf{0.633} &  0.536   \\
    docking   &  0.572 & 0.533  \\
    Ragoza et al.\cite{Ragoza2017}   & - & 0.518  \\
    Torng et al.\cite{Torng2018}   &  - & \textbf{0.563}  \\
    \hline
  \end{tabular}
  \label{tab:Table6}
\end{table}

Additionally, we validated our model on the MUV\cite{Rohrer2009} dataset. The MUV dataset is designed to remove undesirable bias between active molecules and decoys by optimally spreading active molecules in the chemical space of decoys while maintaining the molecular similarities between active-active molecules and active-decoy molecules. In Table \ref{tab:Table6}, our model, as with the other deep learning models and docking, shows AUROC close to 0.5 for the MUV dataset. When a model gives AUROC close to 0.5, it means that the model randomly classifies active and inactive molecules. The considerable performance drop for the ChEMBL and MUV datasets indicates that the deep learning models have common problems with generalization. We hypothesize that the reason for such common performance drop is that molecules available in the DUD-E dataset is simply not enough to cover the vast array of molecules available in the nature. 

\section{Conclusion}
In this work, we proposed a novel approach for representing protein-ligand complexes using a specialized graph neural network suited for predicting drug-target interaction. We directly incorporate the 3D structure of a complex into an adjacency matrix. We also apply a distance-aware graph attention algorithm with gate augmentation to increase the performance of our model. The model was trained using the DUD-E dataset for virtual screening and the PDBbind dataset for pose prediction. As a result, our model outperformed docking and other deep learning based methods in terms of both virtual screening and pose prediction. Our model showed AUROC of 0.968 for virtual screening and 0.935 for pose prediction. It also correctly reproduced the natural population distribution of active and inactive molecules.

Besides the success of our model, it has some limitations in generalization. The performance of the model on classifying experimentally verified active and inactive compounds was significantly reduced compared to the performance on the DUD-E test set. Also, our model, like docking and other deep learning based models, failed to correctly classify active and decoy molecules in the MUV dataset. This may be because the DUD-E training set cannot effectively span huge chemical space.
In addition, because most active molecules share common scaffolds, it is possible that our model may memorize the structural pattern within active molecules instead of recognizing proteins.
Therefore, our next goal is to achieve consistent performance on other independent datasets, such as the MUV and the ChEMBL datasets, as well as the DUD-E test set.

\begin{acknowledgement}

This work was supported by the National Research Foundation of Korea (NRF) grant funded by the Korea government (MSIT)(NRF-2017R1E1A1A01078109).

\end{acknowledgement}




\bibliography{achemso-demo}

\end{document}